\documentclass[runningheads]{llncs}
\usepackage[left=20mm,right = 20mm]{geometry}
\usepackage{cite}
\usepackage{amsmath}
\usepackage{amsfonts}
\usepackage{amssymb}
\usepackage{graphicx}
\usepackage{float}
\usepackage{mathtools}
\usepackage{fancyhdr}
\usepackage{lastpage}
\usepackage{commath}
\usepackage{algpseudocode}
\usepackage[linesnumbered,ruled]{algorithm2e}
\DeclarePairedDelimiter\floor{\lfloor}{\rfloor}
\begin{document}
	\title{SimPatch: A Nearest Neighbor Similarity Match between Image Patches}
	\author
	{Aritra Banerjee}
	\authorrunning{A. Banerjee}
	 \institute{Imperial College London\\
	 aritra306@gmail.com
	 }
	\maketitle
	\begin{abstract}
	 Measuring the similarity between patches in images is a fundamental building block in various tasks. Naturally, the patch-size has a major impact on the matching quality, and on the consequent application performance. We try to use large patches instead of relatively small patches so that each patch contains more information. We use different feature extraction mechanisms to extract the features of each individual image patches which forms a feature matrix and find out the nearest neighbor patches in the image. The nearest patches are calculated using two different nearest neighbor algorithms in this paper for a query patch for a given image and the results have been demonstrated  in this paper.
	\end{abstract}
	\section{Introduction}
	Self-similar patch search is essential for a lot of applications like in image restoration, video compression  based on motion estimation, saliency detection or in painting. Each patch contains the information of that image patch area which can be easily used to complete an incomplete image. If we know the features of a nearby patch we can use it to fill the gaps in images. We have tried to use large patches of size so that it is contains more information in each patch and its easier to match nearby patches which turn when applied to inpainting can fill the nearby image gaps. In this paper we will not only find out the similar image patches but also will compare different algorithms like the Brute Force k-NN algorithm search and KD Tree algorithm \cite{kumar2008good} to define the best algorithm in terms of accuracy and time complexity. The basic idea behind this work is extracting the individual image patch features using some prominent filters like the local binary pattern, Co-occurence matrix and the Gabor filter and then finally use the best search algorithm to find the similar image patches within the image. The results of our work can easily deduce the algorithm which will provide a much accurate and faster results. The different feature extraction techniques produce a feature vector. Combining the feature vector for all the patches forms a feature matrix of $(N_{p}*N_{f})$ dimensions, where $N_{p}$ denotes number of patches in the image and $N_{f}$ denotes the number of points in each feature vector. This matrix can be easily used in the nearest neighbor algorithms to calculate the closest patches.
	\section{Existing Work}
	The existing PatchMatch\cite{Barnes:2009:PAR} algorithm uses a random search to seed the patch matches and iterates for a fewer number of times to iterate good matches. It finds the similarity between two selected patches using a distance function for two different images. So, for a given patch coordinate \textit{a} in image \textit{A} and its corresponding nearest neighbor \textit{b} in image \textit{B}, \textit{f(a)} is actually \textit{$b-a$}. But the PatchMatch algorithm is not as accurate as LSH or kd-trees and increasing its accuracy requires more iterations that cost much more time\cite{korman2011coherency}. The main assumption relies on the incoherency of the image which becomes invalid in some cases such as in strongly textured regions.
	Ruzic and Pizurica\cite{ruvzic2015context} utilize patch matching in the context of image inpainting, finding the best matching patches within the same image. They do this by constraining the search region of the image by looking at the context of patches – their surrounding area. They propose two different ways of defining a context of a patch (one having constant size and the other variable size depending on the image content). The framework that is described in this paper can be used with different context descriptors, but the one chosen here was normalized text on histograms computed from Gabor filter responses. Romano and Elad\cite{romano2016patch} also use context of the patches to enrich the information they contain. Their idea is to have the benefits of using larger patches (21x21 pixels) while working with the smaller amount of data. They concatenate small patches (7x7 pixels) with their context feature – a compact representation of their larger surroundings. The context feature is obtained by measuring the similarity of a small patch to its neighbouring patches, organized as a normalized histogram.
	\section{General Idea of the Employed Feature extraction Techniques}
	Feature extraction reduces a high dimensional space to a space of fewer dimensions\cite{fodor2002survey}. We use different texture analysis\cite{tuceryan1993texture} methods in this paper as descriptors to extract the features from the image patches. Consequently it reduces the dimensions of the image patches in space. We project each patch in the feature vector space which reduces the dimensions. Hence, from the feature space we can also find out the nearest neighbors as well. To better demonstrate the concept of feature space we extract the features of each patch and project it in space.
	Let us take a sample image.
		\begin{figure}[H]
			\centering
			\includegraphics[scale = 0.2]{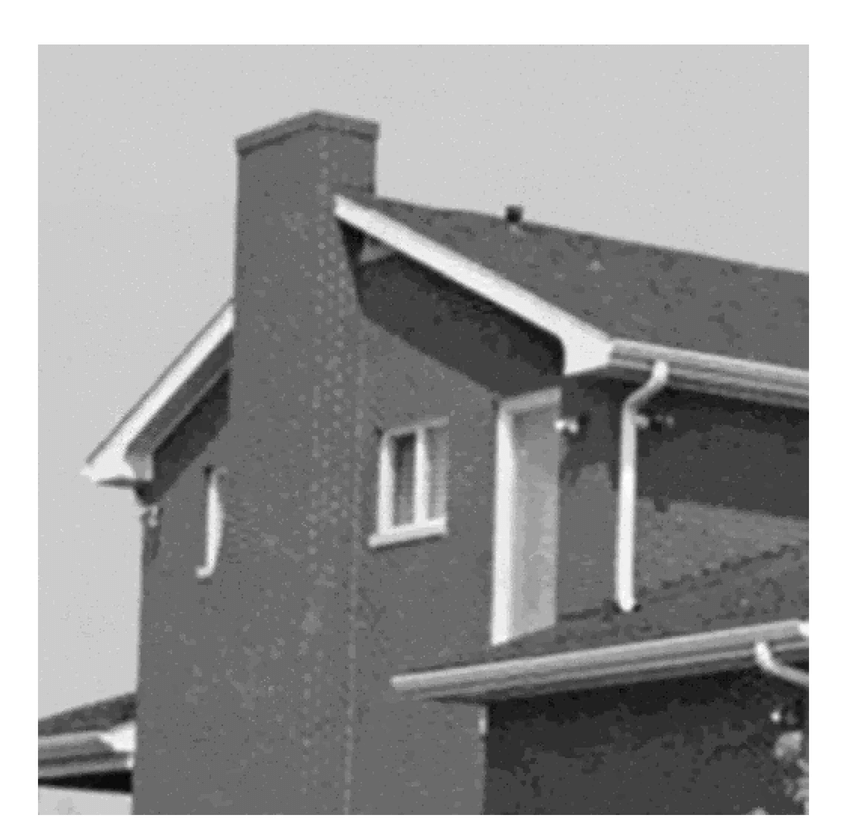}
			\caption{A sample Image}
			\label{fig:house}
		\end{figure}
	Now from this image A we take random patches which are similar to each other from a viewer's perspective and visualize the feature vector space. The nearest patches are projected near each other in the feature vector space. This makes it easier to calculate the nearest patches using the distance metric.
	\begin{figure}[H]
		\centering
		\includegraphics[scale = 0.5]{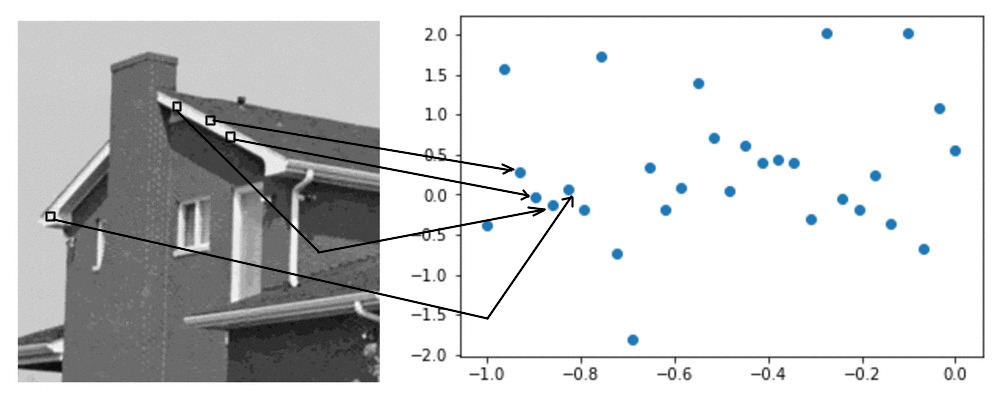}
		\caption{Similar patches get projected nearby in the feature space}
		\label{fig:fspace}
	\end{figure}
	As we can see in Fig. \ref{fig:fspace} the similar patches irrespective of their size and lcoation in the image itself get projected near each other in the vector space. This works similarly for all patches in the rest of the image as well. The size of the patch do not change the fact that nearby patches get projected to the nearby positions in vector space. In our paper we have used the techniques used mostly in texture analysis to extract the features like the Local Binary Pattern\cite{he1990texture}, Gray Level Co-occurence matrix and Gabor Filters\cite{daugman1988complete}.
	\section{Techniques employed for feature extraction}
	Extraction of image patch features is an essential part to deciding the nearest neighbors. We extract the features and necessarily store it in a feature vector which in turn provides us the basic similarity score when we apply the nearest neighbor algorithms on the feature vectors. The feature vectors contain a lot of information about the image patches itself.
	\subsection{Local Binary Pattern}
	Esfahanian, M et al. \cite{esfahanian2013using} used a technique called the local binary pattern to generate feature vectors from images for dophin vocalization classification. The local binary pattern\cite{ojala1996comparative} or the LBP operator is used as a descriptor for local spatial patterns and the gray contrast. Fundamentally an LBP label is a binary digit which is assigned to each of the bits in the patch based on its difference from one of the pixels for a given value of radius. Now let us look at the algorithm in detail. We generally use a grayscale patch for this case. If an image is not in grayscale, first we convert it into grayscale and then apply LBP. For each pixel in the grayscale patch, we select a neighborhood of size \textit{'r'} surrounding the center pixel. An LBP value is then calculated for the center pixel and stored in the output 2D array which has the same width and height as the input patch. This provides us with a binary number for each computation. Consequently we can show the above explanation in terms of a mathematical formula for the value of the LBP code of a pixel $(x_{c},y_{c})$:
	
	\begin{equation}
	LBP_{P,R} = \sum_{p=0}^{P-1} s(g_{p}-g_{c})2^{P}
	\label{eq:1}
	\end{equation}
	where 
	
	\begin{equation}
	s_{x}=\begin{cases}
	1, & \text{if $x>=0;$}\\
	0, & \text{otherwise}.
	\end{cases}
	\label{eq:2}
	\end{equation}
	After getting the binary number we compute the histogram, over the cell, of the frequency of each "number" occurring(i.e., each combination of which pixels are smaller and which are greater than the center). This is typically a 256 dimensional feature vector.We can then normalize the histogram and concatenate histograms of all cells. This gives a feature vector for the entire window of the patch. In the proposed algorithm we use two metrics from LBP features namely:
	\begin{itemize}
		\item Energy
		\item Entropy
	\end{itemize}
	Now we take an example image and try to find out the local binary pattern of the image and also the LBP energy and LBP entropy.
	We will consider one single image and find out different features for that particular image step by step.
	\begin{figure}[H]
		\centering
		\includegraphics[scale = 0.5]{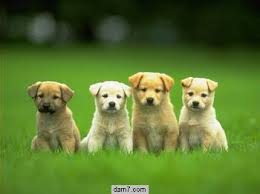}
		\caption{The input image for calculating features}
		\label{fig:images}
	\end{figure}
	We use LBP on Fig. \ref{fig:images} using the formula stated in Eqns. \eqref{eq:1} and \eqref{eq:2} to get the image displayed below:
	\begin{figure}[H]
		\centering
		\includegraphics[scale = 0.5]{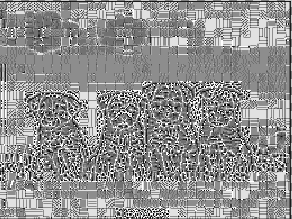}
		\caption{Image after applying LBP}
	\end{figure}
	After applying LBP we calculate the LBP entropy and the LBP energy from an 8 bin histogram in this case. We get LBP energy = \textbf{0.189} and LBP entropy = \textbf{2.647}.
	\subsection{Gray Level Co-occurence Matrix}
	Pathak et al. \cite{pathak2013texture} used gray level co-occurence matrix on an 8 bit gray scale image to describe the surface texture and derive second order statistical information of neighboring pixels of an image. Sebastian et al. \cite{sebastian2012gray} defined a new feature trace which was extracted from the GLCM(Gray Level Co-occurence Matrix) and how it is implemented in texture analysis with context to Context Based Image Retrieval(CBIR). Technically it is a matrix that is defined over an image patch in our case to be the distribution of co-occuring grayscale pixel values at a particular given offset. Before explaining this approach let us need to understand a few concepts:
	\begin{itemize}
		\item The offset$(\Delta a,\Delta b)$ is defined which is the position operator which is relative to an pixel of the image.
		\item For the image patch with \textit{p} pixel values  a \textit{p\textnormal{x}p} co-occurence matrix is generated.
		\item The $(x,y)^{th}$ value of the co-occurence matrix gives the number of times in the patch that the $x^{th}$ and $y^{th}$ pixel values occurs.
	\end{itemize}
	Hence, for an image with \textit{p} different pixel values, the \textit{p\textnormal{x}p} co-occurence matrix \textbf{M} is defined for an \textit{n\textnormal{x}m} patch with the parameters $(\Delta a,\Delta b)$ as offset:
	
	\begin{equation}
	M_{\Delta a,\Delta b}(x,y)=\sum_{a=1}^{n}\sum_{b=1}^{m}\begin{cases}
	1, & \text{if $I(a,b)=x$,}\\
	   & \text{$I(a+\Delta a,b+\Delta b)$}\\
	   & \text{$=y;$}\\
	0, & \text{otherwise}.
	\end{cases}
	\label{eq:3}
	\end{equation}
	where, \textit{x} and \textit{y} are pixel values, \textit{a} and \textit{b} are spatial positions in image \textbf{I}; and I(a,b) indicates the pixel value at pixel (a,b).\\Since the co-occurrence matrices are typically large and sparse, various metrics of the matrix are often taken to get a more useful set of features. In the proposed algorithm we use five metrics from it namely: \begin{itemize}
		\item Contrast
		\item Dissimilarity
		\item Homogeneity
		\item Energy
		\item Correlation
	\end{itemize} Features generated using this technique are usually called Haralick features, which was named after Robert Haralick\cite{haralick1973textural} after his contributions to image classification.
	For GLCM we use Fig. \ref{fig:images} on Page \pageref{fig:images} to calculate the five metrics specified above to add to our feature vector using the Eqn. \eqref{eq:3}. First we calculate the Co-occurence matrix and then from the Co-occurence matrix we caluclate the values of Contrast = \textbf{277.319}, Dissimilarity = \textbf{7.178}, Homogeneity = \textbf{0.413}, Energy = \textbf{0.036} and Correlation = \textbf{0.948}.
	\subsection{Gabor Filter}
	Roslan et al.\cite{roslan2012texture} used 2D Gabor filters as a method to extract texture features of Inverse Fast Fourier transform(IFFT),texture energy and transformed IFFT for an image. We have used a gabor filter with a certain set frequency and derived the real part and imaginary part of the filter which is followed by the magnitude function which helps us get only the real part of the gabor filter that is ultimately used to compute the gabor histogram and probability. Again we have used two metrics of Gabor filter to add to our feature vector:
	\begin{itemize}
		\item Gabor energy
		\item Gabor entropy
	\end{itemize}
	We can represent the real and imaginary part of the Gabor filter using the following mathematical formula which represent the orthogonal directions:
	\begin{itemize}
		\item Real
		\begin{equation}
		\begin{split}
		g(x,y;\lambda ,\theta ,\psi ,\sigma ,\gamma) = exp\bigg(-\dfrac{x'^{2}+\gamma ^{2}y'^{2}}{2\sigma ^{2}}\bigg)\\ \cos \bigg(2\pi \frac{x'}{\lambda}+\psi \bigg)
		\end{split}
		\label{eq:4}
		\end{equation}
		\item Imaginary
		\begin{equation}
		\begin{split}
		g(x,y;\lambda ,\theta ,\psi ,\sigma ,\gamma) = exp\bigg(-\dfrac{x'^{2}+\gamma ^{2}y'^{2}}{2\sigma ^{2}}\bigg)\\ \sin \bigg(2\pi \frac{x'}{\lambda}+\psi \bigg)
		\end{split}
		\label{eq:5}
		\end{equation}
	\end{itemize}
	where,
	
	\begin{equation}
	x' = x\cos \theta + y\sin \theta
	\label{eq:6}
	\end{equation}
	and
	
	\begin{equation}
	y' = -x\sin \theta + y\cos \theta
	\label{eq:7}
	\end{equation}
	Here, $\lambda$ represents the wavelength of the sinusoidal factor, $\theta$ represents the orientation of the normal to the parallel stripes of a Gabor function\cite{gabor1946theory}, $\psi$ is the phase offset, $\sigma$ is the standard deviation of the Gaussian envelope and $\gamma$ is the spatial aspect ratio.
	Using Eqns. \eqref{eq:4},\eqref{eq:5},\eqref{eq:6} and \eqref{eq:7} we calculate the real and imaginary part of the Gabor filter for Fig. \ref{fig:images} on Page \pageref{fig:images}.
	\begin{figure}[H]
		\begin{minipage}{0.3\textwidth}
		\includegraphics[width=\linewidth]{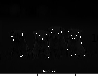}
		\caption{The Real Part of Gabor Filter}\label{fig:GaborR}
		\end{minipage}\hfill
		\begin{minipage}{0.3\textwidth}
		\includegraphics[width=\linewidth]{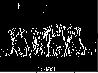}
		\caption{Imaginary Part of Gabor Filter}\label{fig:GaborI}
		\end{minipage}\hfill
		\begin{minipage}{0.3\textwidth}%
		\includegraphics[width=\linewidth]{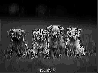}
		\caption{Magnitude of Gabor Filter}\label{fig:GaborM}
		\end{minipage}
	\end{figure}
	As we can observe that the Fig. \ref{fig:GaborR} denotes the real part, Fig. \ref{fig:GaborI} denotes the imaginary part and Fig. \ref{fig:GaborM} denotes the magnitude part of the gabor filter. Ultimately we need to work with the magnitude in this case because we need to find the feature metrics. From the magnitude of the gabor filter using an 8 bin histogram we calculate the energy and entropy for gabor filter. After calculation for the Figure \ref{fig:images} we get Gabor Energy = \textbf{0.291} and Gabor Entropy = \textbf{2.084}
	\section{Methods used for finding Nearest Neighbors}
	In this section we clearly describe the methods we have used in this paper to find out the nearest patches in an image. Mostly we have used euclidean or cosine distances to find nearest patches for different methods but other distance measures can also be used.
	\subsection{Brute Force k-Nearest Neighbor Algorithm(Using Cosine Similarity)}
	The K-nearest neighbor algorithm\cite{dasarathy1991nearest} is one of the fundamental searches that is used in pattern recognition. In this paper we deal with the k-means classification in the nearest neighbor feature space.Wettschereck et al.\cite{wettschereck1994locally} used different methods to compare with the basic k-nearest neighbor algorithm using a locally adaptive nearest neighbor algorithm. Although in our algorithm we only use the basic k-nearest algorithm. The distance metric is taken as cosine distance. Anna Huang\cite{huang2008similarity} worked with different text document for clustering and similarity checking and used cosine distance as one of the better methods to be used as a distance metric. Hence, we used cosine distance as a metric while applying nearest neighbor search. Baoli Li et al.\cite{li2013distance} used cosine similarity as the basic component in text mining applications and have proved that more accurate results can be expected in cosine similarity than other distance metrics. The cosine similarity formula can be mathematically shown as below:
	
	\begin{equation}
	cos(\theta) = \frac{\textbf{AB}}{\norm{\textbf{A}}\norm{\textbf{B}}}
	\end{equation}
	where, A and B are two feature vectors in this scenario.
	The resulting similarity ranges from −1 meaning exactly opposite, to 1 meaning exactly the same, with 0 indicating orthogonality (decorrelation), and in-between values indicating intermediate similarity or dissimilarity.
	\subsection{kd-Trees Algorithm(Using Euclidean Similarity)}
	Bentley\cite{bentley1975multidimensional} applied the kd-tree algorithm to retrieve information by associative searches. It is basically a multidimensional binary tree search, where \textit{k} denotes the dimensionality of the search space. The drawing of the tree is quite simple. At each node we recursively partition the node into two sets by splitting along one dimension of the data, until one of the terminating criteria is met with. Basically while calculating we need to choose two things: the split dimension and the split value. The split dimension is chosen as such that the dimension with maximum variance leads to smaller trees\cite{bentley1990k} and the split value is generally chosen as the median\cite{kumar2008good} along the split dimension. The reason for using kd tree is that the running time for nearst neighbor search is $O(log N)$\cite{bentley1975multidimensional} which helps us reduce the running time of high dimensional datasets\cite{marimont1979nearest}. Among the valid distance metric\cite{wu2011randomly} that can be used we have chosen euclidean distance to calculate the similar vectors.
	\section{Designed Algorithm for Matching Similar Patches}
	The concept we have used in our paper is a combination of image processing and machine learning. Firstly, we extract patches from an image, store the patches and generator feature vectors for each patch and then using nearest neighbors algorithm find the nearest neighbors of the desired patch as stated by the user with the help taken from feature vectors. The image patch size and also the number of k-nearest neighbors to be located is typically provided by the user. We will use both the methods to extract the patches separately and then compare the speed of each algorithm. The patches extracted are first displayed to the user and then they are plotted on the original image for better visualization.\\
	The algorithm is stated as \textbf{Algorithm \ref{alg:algo1}}.
		\begin{algorithm}[!ht]
		\caption{SimPatch}
		\label{alg:algo1}
		\SetAlgoLined
		\DontPrintSemicolon
		\SetKwBlock{Begin}{Begin}{End}
		\Begin{
		\KwIn{Image A$[M*N]$, Individual Patch Size and No. of nearest neighbors(k)}
		\KwOut{Nearest patches of given patch size and Mapped Nearest patches on original image A}
		Convert the image A to gray level image, extract the 2D patches from the image on the basis of individual patch size and store it in avariable \textit{temp}.\;
		$i \gets 0$\;
		$no. of patches \gets length(temp)$\;
		\While{$i<=no. of patches$}
		{
			Apply feature extraction techniques employed for each patch,  concatenate the values of each patch generated from each feature technique and form a feature matrix \textit{feat} of dimensions ($no. of patches*no. of features$).
		}
		Normalize the values in the feature matrix $(feat \in [0....1])$. \;
		Use left mouse click in event handling to display an image and return the coordinates as \textit{coords} of a patch when left clicked on the image. \;
		$patch no. = coords[x-coordinate]*N + coords[y-coordinate]$\;
		Calculate the nearest patches by the specified nearest neighbor algorithm(k-nearest or kd-trees) using $patch no., k$(number of neighbors) and the distance metric (cosine or euclidean) and store it in a variable \textit{ind}. \;
		Display the nearest patches extracted from each method (k-nearest and kd-trees). \;
		$j \gets 0$\;
		\While{$j<=k$}
		{
			$t \gets ind[j]$ \;
			$xcoord \gets \floor*{\frac{t}{N}}$ \;
			$ycoord \gets t - (xcoord*N)$ \;
			Draw a rectangle for each patch coordinates ($xcoord,ycoord$) generated on the image using the appropriate individual patch size. \;
		}
		Display the original image A with all the nearest patches marked on it in form of rectangles. \;
	}
		\end{algorithm}
	\section{SimPatch Implementation}
	In this section we show the SimPatch algorithm execution on Fig. \ref{fig:images} on Page \pageref{fig:images} using the two different nearest neighbor methods mentioned, each of these methods which uses all the three feature extraction techniques specified. For Fig. \ref{fig:images} we get a total of \textbf{44426} patches if the patch size is $(32*32)$. The patch size is specified by the user. So after calculating the feature vectors for each of the 44426 patches, it returns a $(44426*9)$ feature matrix. Then after clicking on a random point on Fig. \ref{fig:images}, the patch coordinate is returned as $(\textbf{113,104})$ as illustrated in Fig. \ref{fig:location} below:
	\begin{figure}[H]
		\centering
		\includegraphics[scale = 0.35]{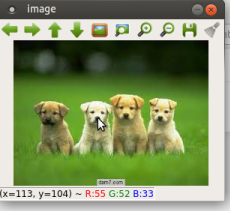}
		\caption{Query Patch selected from image}
		\label{fig:location}
	\end{figure}
	Now from the coordinates returned we calculate the query patch using the formula given in line 13  in Algorithm \ref{alg:algo1}. Then we input the number of neighbors k = \textbf{5}. After this we calculate the \textit{k} nearest neigbors for the image.
	\subsection{Using Brute Force k-Nearest Neighbor}
	This method typically calculates the nearest neighbors in $O(N)$\cite{verma2014comparison} time complexity. It returns the five(\textit{k}) nearest patch IDs based on cosine distance. Then we display the nearest patches as:
	\begin{figure}[H]
		\centering
		\includegraphics[scale = 0.3]{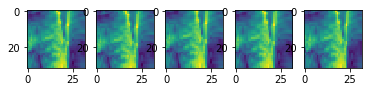}
		\caption{The five nearest patches extracted}
		\label{fig:brutepatches}
	\end{figure}
	Now for visualization we display the patches on the input image by drawing rectangles on patches with the patch size $(32*32)$ as dimensions:
	\begin{figure}[H]
		\centering
		\includegraphics[scale = 0.4]{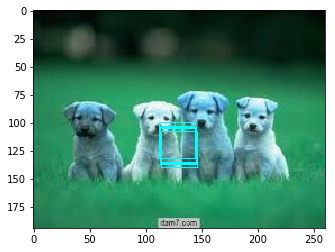}
		\caption{Five nearest patches displayed from query patch}
		\label{fig:bruteimage}
	\end{figure}
	For the computation of this nearest neighbor module it took approximately $2.3287 seconds$.
	\subsection{Using kd-trees for k-Nearest Neighbor}
	Again,we try to detect the nearest image patches for the same image and for the same query patch in that image using the kd-trees method which calculates the nearest neighbors in $O(log N)$ time complexity. It returns the five(\textit{k}) nearest patch IDs based on euclidean distance. Notice that leaving aside only the last patch it returns the same nearest neighbor matches for the remaining four patches. Then we display the nearest patches as:
	\begin{figure}[H]
		\centering
		\includegraphics[scale = 0.3]{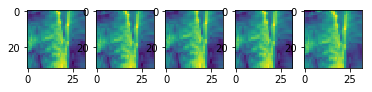}
		\caption{The five nearest patches extracted}
		\label{fig:kdpatches}
	\end{figure}
	Again, for visualization we display the patches on the input image by drawing rectangles on patches with the patch size $(32*32)$ as dimensions:
	\begin{figure}[H]
		\centering
		\includegraphics[scale = 0.4]{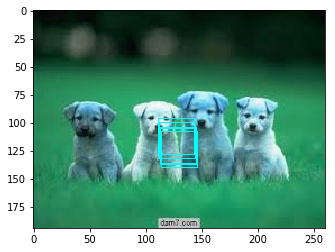}
		\caption{Five nearest patches displayed from query patch}
		\label{fig:kdimage}
	\end{figure}
	In this case, the computation of this nearest neighbor module it took approximately $0.0309 seconds$.
	\subsection{Graphical Comparison between two methods}
	In this section we will explain graphically the comparison between Brute Force k-NN and kd-Trees method. We will plot a distance and time graph which will help us calculate the maximum speed of each algorithm. The graph is represented as below:
	\begin{figure}[H]
		\centering
		\includegraphics[scale = 0.5]{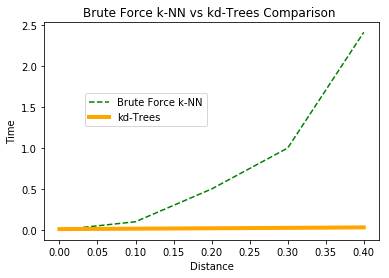}
		\caption{Distance vs Time Comparison between Brute Force k-NN and kd-Trees}
		\label{fig:comparison}
	\end{figure}
	So, from the graph for Brute Force k-NN the maximum time($t_{max}$) = $2.3287$ and the maximum distance($d_{max}$) = $0.4$. Hence, the maximum speed($s_{max}$) = $\textbf{0.1717}units$ approximately.
	And, for kd-trees the maximum time from graph($t_{kdmax}$) = $0.0309$ anf the maximum distance = ($d_{kdmax}$) = $0.4$. Hence, the maximum speed($s_{kdmax}$) = $\textbf{12.9449}units$ approximately.
	As we clearly observe that the speed for computation is around \textbf{12} times more in the case of kd-Trees than in the case of Brute Force k-NN. Therefore, it can be concluded that the kd-Trees method provide the better computation cost as compared to brute force k-NN while finding image patches and the accuracy is nearly unaffected.
	\section{Conclusion}
	In this paper we used different feature extraction techniques to apply to all the patches of the image and form a feature vector of the entire patch dataset. From the feature matrix generated we used different nearest neighbor methods to calculate similar patches in the image. This process can be further refined by using other descriptors like LSH\cite{kulis2009kernelized}, SIFT\cite{lowe2004distinctive} descriptor or probably using Convolutional Neural Networks (CNN)\cite{melekhov2016image}. Using all these other descriptors we can compare with the SimPatch algorithm that we presented in this paper.
	\bibliographystyle{splncs04}
	\bibliography{references1}
\end{document}